# An Enhanced Indexing And Ranking Technique On The Semantic Web

Ahmed Tolba[1], Nabila Eladawi[2] and Mohammed Elmogy[3]

[1] Faculty of Computer Studies, Arab Open University
Kuwait, 3322, Kuwait

[2] Faculty of Computers and Information, Mansoura University
Mansoura, 35516, Egypt

[3] Faculty of Computers and Information, Mansoura University
Mansoura, 35516, Egypt

**Abstract**
With the fast growth of the Internet, more and more information is available on the Web. The Semantic Web has many features which cannot be handled by using the traditional search engines. It extracts metadata for each discovered Web documents in RDF or OWL formats, and computes relations between documents. We proposed a hybrid indexing and ranking technique for the Semantic Web which finds relevant documents and computes the similarity among a set of documents. First, it returns with the most related document from the repository of Semantic Web Documents (SWDs) by using a modified version of the ObjectRank technique. Then, it creates a sub-graph for the most related SWDs. Finally, It returns the hubs and authorities of these document by using the HITS algorithm. Our technique increases the quality of the results and decreases the execution time of processing the user's query.

## 1. Introduction

The classical Information Retrieval (IR) models have been processed by using state-of-the-art models such as LSI and machine learning based models (i.e. artificial neural network, symbolic learning, and genetic algorithm) [1]. However, it has been shown that these models based on formal mathematical theories and they do not necessarily surpass the classical models [2]. In the classical IR models, matching between queries and documents is formally defined, but it is semantically imprecise. Most of these models make a plausible assumption that words in documents are independent.

On the other hand, human users are able to interpret the significance of semantic features to understand the information being presented, but this may not be so easy for an automated process or software agent. The Semantic Web aims to overcome this problem by making Web content more accessible to automated processes. The ultimate goal of the Semantic Web is to transform the existing Web into a set of connected applications and forming a consistent logical Web of data [3,4]. This can be achieved by adding semantic annotations that describe the meaning of the Web content.

Therefore, the Semantic Web will contain resources corresponding not only to media objects (such as Web pages, images, audio clips, etc.) as the current Web does, but also to objects such as people, places, organizations, and events [5]. Consequently, the Semantic Web will contain not just a single kind of relation (the hyperlink) between resources, but many different kinds of relations between the different types of resources.

This paper is divided into five sections. In Section 2, an overview of the related work, which discusses some recent ranking systems on the Semantic Web, will be introduced. Section 3 represents the architecture of our proposed system. The proposed indexing and ranking technique is described in detail. Section 4 presents the implementation and some results of our system. Finally, we conclude our work in Section 5.

## 2. Related Work

There are many researchers who are working on Semantic Web and how to rank the pages according to their contents. For example, TAP [6,7] was created to be an infrastructure for applications on the Semantic Web. It provides a set of simple mechanisms for sites to publish data onto the Semantic Web and for applications to consume this data. TAP improves information search and retrieval results in two ways: on the one hand, it provides a simple mechanism to help the Semantic search module to understand the denotation of the query; on the other hand, it augments the search results by considering search context and exploring closely related objects based on this context.

Kiryakov et al. [8] introduced a holistic architecture of Semantic annotation, indexing, and retrieval for documents. Their system, which is called KIM, aimed to achieve fully automatic annotation and to improve search and retrieval by integrating information extraction (IE) (i.e. using GATE [9]), information retrieval and Semantic Web technologies.

In [10-12], the authors viewed the documents representation on the Semantic Web as a combination of text, which is suitable for current Web search engines' indexing and Semantic markup. This can be used to perform inference over a knowledge-base and proposes an integrated approach to combine the inference capability and traditional information retrieval techniques. They implemented a prototype system, called OWLIR, for retrieving university event announcements.

Squiggle [13] is another framework for building domain-specific Semantic search applications. It provides capabilities for annotating, indexing, and retrieving multimedia items based upon the SKOS3 ontology.

Swoogle [14,15] is also a Semantic search engine for retrieving Semantic Web document. Its primary use is found in searching the Web and locating relevant ontologies in order to help users access, explore, and query Semantic Web documents.

Stojanovic et al. [16] have developed a domain independent approach for developing Semantic portals, viz. SEAL (SEmantic portAL), that exploits Semantics for providing and accessing information at a portal as well as constructing and maintaining the portals. They propose that the problem of Semantic ranking may be reduced to the comparison of two knowledge-bases. Query results are reinterpreted as "query knowledge bases" and their similarity to the original knowledge-base without axioms yields the basis for Semantic ranking. Thereby, they reduce their notion of similarity between two knowledge bases to the similarity of concept pairs.

Yousefipour et al. proposed an ontology-based approach for ranking Semantic Web services. A generic and domain-specific ontology is used to infer the Semantic similarity between the parameters of the request and the advertisement, which will be applied in the process of SWSs ranking. They studied how Semantic Web service ranking can be used in the context of Semantic Web service discovery.

Therefore, there are many researches on Semantic Web and how to rank the pages according to their contents. As mentioned previously, these ranking techniques do not depend only on the keywords but also on the contents of the Web documents. Consequently, the Semantic Web ranking techniques need to be developed to present an efficient way to classify SWDs and to retrieve a precise result for the user's query.

We developed an indexing and ranking technique which can be used in a Semantic Web search engine to facilitate the development of the Semantic Web and finding a proper ontology for the submitted search query. The entire documents are processed to extracted ontologies and find the relationships between these documents and the others. Therefore, our system is not based only on the extracted metadata from the document but also on extracted ontologies and the relations between documents. In other words, our main goals are to find a good measure for indexing the processed SWDs with extracting the proper ontologies, create a meaningful rank measure which reflects the importance of the processed document, and answer the user's queries efficiently.

## 3. System Architecture

The main architecture of our system is as shown in Fig. 1. Our system contains five components: (1) The JENA Web crawler which is used to crawl the Web and returns with SWDs to process them, (2) SWDs and metadata repositories which contain the retrieved Semantic Web Documents and their extracted metadata, (3) The Semantic ranking component which is used to index and rank the processed SWDs, (4) The pre-processing stage which stores SWDs in the repository and processes the SWDs to generate objective metadata about SWDs at both the syntax and the semantic levels, and (5) The user interface which accepts the query from the user and displays the result of the search. In the following subsections, we will discuss the components of our system in more detail.

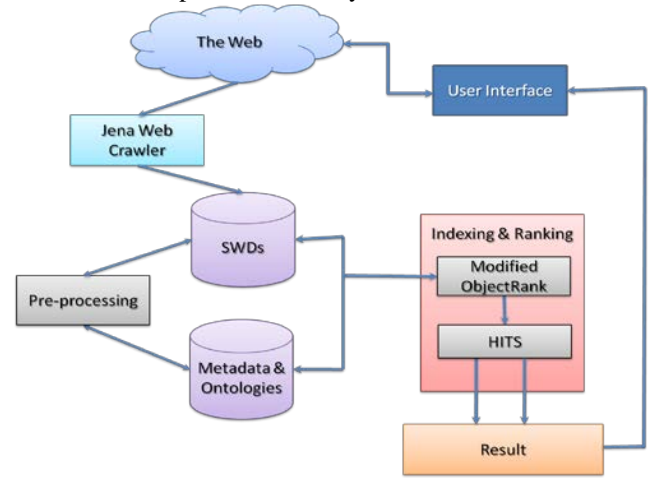

Fig. 1 The architecture of the proposed system.

### 3.1 Jena

Jena [18] is a Java framework for building Semantic Web applications. It provides a programmatic environment for RDF, RDFS and OWL, and SPARQL. It includes a rule-based inference engine. It is open source and grown out of work with the HP Labs Semantic Web Programme. Jena is developed to employ a number of heuristics for finding SWDs. It searches for documents of .rdf, .owl, .daml, and .n3 file extensions.

Jena analyzes the content of a SWD and discovers new SWDs. First, it verifies if a document is a SWD or not, and it also revisits discovered URLs to check updates. Secondly, several heuristics are used to discover new SWDs through semantic relations: (1) The semantics of URIref shows that the namespace of a URIref is highly likely to be the URL of an SWD; (2) The semantics of OWL shows that owl:imports links to an external ontology, which is a SWD; (3) The semantics of FOAF ontology, shows that rdfs:seeAlso property of an instance of foaf:Person often links to another FOAF document, which often is a SWD.

### 3.2 SWDs Repository

SWDs repository contains the Semantic Web documents which are retrieved by the Web crawler. It keeps up-to-date SWDs to use them in to answer user's query. SWDs are based on RDF which can be in RDFS, DAML+OIL, or OWL formats. They contain the following items:

- General term statements which define the classes and the properties.
- The terms' definition extensions.
- Individuals Creation.
- Make assertions about terms and individuals which are already defined or created.

Therefore, SWD can be defined as an atomic information exchange object in the Semantic Web which can be found online and accessible to Web users and software agents.

### 3.3 SWDs Pre-processing

The stored SWDs in the repository are processed to generate objective metadata about SWDs at both the syntax and the semantic levels. The SWDs are classified in the repository into three types: the Semantic Web ontologies (SWOs) which is called T-Boxes, the Semantic Web databases (SWDBs) which is called A-Boxes, and hybrid which defines a set of terms to be used by others as well as a useful database of information about a set of individuals.

SWD metadata is collected to make SWD processing and search more efficient and effective. SWD metadata classification can considered as a modified version of the one which is used in Swoogle. We added some additional items and changed others. Fig. 2 shows the types of SWDs metadata which are processed and stored in metadata and ontologies repository.

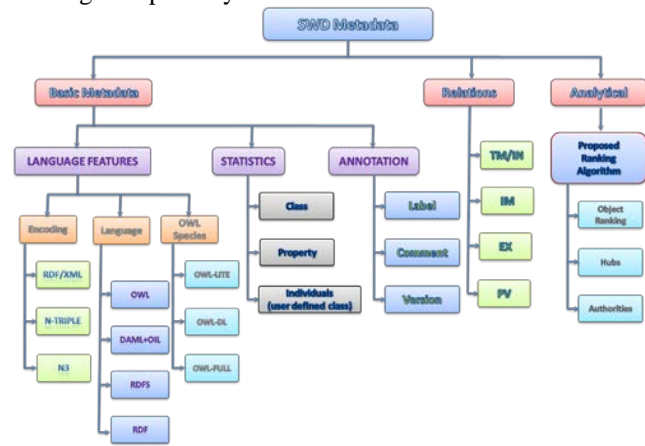

Fig. 2 Types of SWDs metadata stored in metadata &ontologies repository.

SWD metadata is derived from the content of SWD as well as the relations among SWDs. They can be classified into three categories of metadata:

- **Basic Metadata**: It considers the syntactic and semantic features of a SWD. It contains the following types:
    - Language feature: It refers to the properties describing the syntactic or semantic features of a SWD. It captures the following features:
        - Encoding: It shows the syntactic encoding of a SWD.
        - Language: It shows the language used by a SWD.
        - OWL Species: It shows the language species of a SWD written in OWL.
    - RDF Statistics: It refers to the properties summarizing node distribution of the RDF graph of a SWD.
    - Ontology annotation: It refers to the properties that describe a SWD as an ontology.
- **Relations:** They consider the explicit semantics between individual SWDs. Table I shows the different types of relations which can classified into four categories:
    - TM/IN captures term reference relations between two SWDs.
    - IM shows that an ontology imports another ontology.
    - EX shows that an ontology extends another.
    - PV shows that an ontology is a prior version of another.

- **Analytical Metadata**: It contains the SWO/SWDB classification and SWD ranking. The proposed indexing and ranking technique will discuss in the following subsection.

Table 1: The Types of the relationships among SWDs.

| Type | Classes and Properties |
|---|---|
| TM/IN | owl:termRef, daml:termRef |
| IM | owl:imports, daml:imports |
| EX | rdfs:subClassOff, rdfs:subPropertyOf, owl:disjointWith, owl:equivalentClass, owl:equivalentProperty, owl:complementOf, owl:inverseOf, owl:intersectionOf, owl:unionOf, daml:sameClassAs, daml:samePropertyAs, daml:inverseOf, daml:disjoinWith, daml:complementOf, daml:unionOf, daml:disjoinUnionOf, daml:ntersectionOf |
| PV | owl:priorVersion, owl:DeprecatedProperty, owl:DeprecatedClass, owl:backwardCompatibleWith, owl:incompatibleWith |

3.4 Indexing and Ranking Stage

We propose a general approach for Semantic ranking to provide high quality, high recall search in databases and on the Web. A hybrid page ranking technique is proposed which integrate the strength of both ObjectRank [19] which is calculated offline and the Hits [20] search which is run online. Therefore, our hybrid approach is using a number of relatively small subsets of the data graph in such a way that any keyword query can be answered by high ranked documents with only one of the sub-graphs. Our proposed approach tries to find the trade-off between query execution time and quality of the results.

Our technique is divided into two portions: pre-processing and query-time stages. At pre-processing stage, we will apply a modified version of the ObjectRank technique and HITS technique will be applied at the query time. We proposed a combination of these two techniques to avoid the pitfalls of each technique. We also want to benefit from the advantages of both of them. We will discuss these pitfalls and advantages in the following sub-sections.

ObjectRank inspired by the idea of PageRank [21] technique. These algorithms that use PageRank require a query-time PageRank-style iterative computation over the full graph. This computation is too expensive for large graphs, and not feasible at query time, as it requires multiple iterations over all nodes and links of the entire database graph.

On the other hand, one advantage of the HITS algorithm is its dual rankings. HITS presents two ranked lists to the user: one with the most authoritative documents related to the query and the other with the most "hubby" documents.

Authoritative pages relevant to the initial query should not only have large in-degree; since they are all authorities on a common topic, there should also be considerable overlap in the sets of pages that point to them. Thus, in addition to highly authoritative pages, the researchers expect to find what could be called hub pages: these are pages that have links to multiple relevant authoritative pages. It is these hub pages that "pull together" authorities on a common topic, and allow us to throw out unrelated pages of large in-degree. A good hub is a page that points to many good authorities; a good authority is a page that is pointed to by many good hubs

To solve the problems of the ObjectRank and HITS techniques, we join these techniques as follows. First, we calculate the ranking scores for all the SWDs in our database using a modified version of the ObjectRank algorithm save them in a repository indexed with keywords. ObjectRank gives the same initial values for all nodes. In our experiment, we initialized each node with the ratio of all links that the node receives as in-links instead of giving the same initial value for all the pages. The ratio offers an enhanced initial guess with minimal overhead. In experimental evaluation, we found that this initial hypothesis reduces the number of iterations required by about one third. When a user type a query the technique will work as follow:

- We search in the repository for the most n ranked SWDs. this will reduce the time ObjectRank need to look at the entire database for a very large number of output.
- Make a sub-graph around these SWDs by adding the in-links and out-links SWDs.
- for each page in the sub-graph add only d pages from the pages that point to it (in-links), then add all pages that this page point to (out-links).
- Calculate the hub score and authority score for each page in the sub-graph.
- Output the most authoritative pages and the most hubby pages to the user.

The PageRank algorithm evaluates the Web documents according to the following equation:

$$PR(A) = (1-d) + d(PR(T_1)/C(T_1) + PR(T_2)/C(T_2) + ...... + PR(T_n)/C(T_n))$$

where A is a document. $T_1$; $T_2$; .... ; $T_n$ are Web documents that link to A; $C(T_i)$ is the total out-links of $T_i$; and d is a damping factor, which is typically set to 0.85. This equation captures the probability that a user will arrive at a given page either by directly addressing it, or by following one of the links pointing to it.

Unfortunately, this random model is not appropriate for the Semantic Web. Because there is different types of link between SWDs. Therefore, this leads to a non-uniform probability of following a particular outgoing link, So

ObjectRank uses a model which accounts for the various types of links that can exist between SWDs.

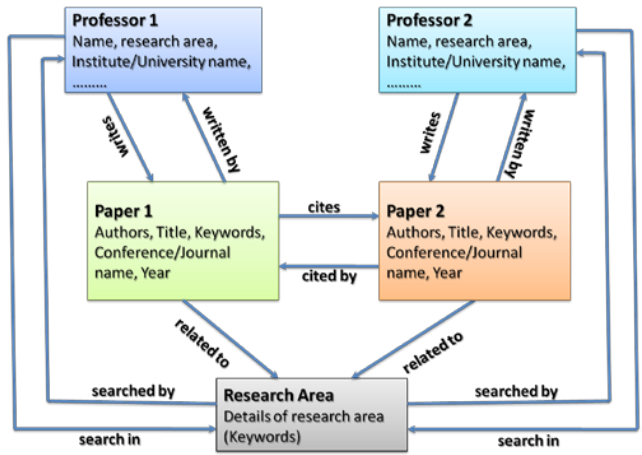

Fig. 3 An example of a subset of the ObjectRank graph.

Consequently, ObjectRank is developed as a system to perform authority-based keyword search on databases which is inspired by PageRank. Fig. 3 shows an example of a subset of ObjectRank graph. The ObjectRank algorithm applies authority-based ranking to keyword search in databases modeled as labeled graphs. Conceptually, authority originates at the nodes (objects) containing the keywords and flows to objects according to their semantic connections. Each node is ranked according to its authority with respect to the particular keywords. One can adjust the weight of global importance, the weight of each keyword of the query, the importance of a result actually containing the keywords versus being referenced by nodes containing them, and the volume of authority flow via each type of semantic connection. This algorithm as we can see is divided into two parts the preprocessing time and this what we are concerned about, and we will modify the query time stage of the ObjectRank technique.

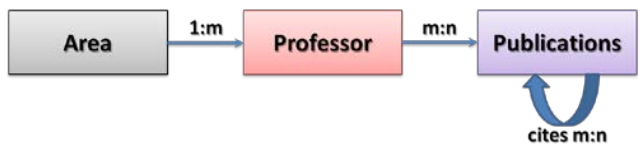

Fig. 4 The ObjectRank schema graph.

They view a database as a labeled graph, which is a model that easily captures both relational and XML databases. The data graph $D(V_D, E_D)$ is a labeled directed graph where every node v has a label $\lambda(v)$ and a set of keywords. Each node represents an object of the database and may have a sub-structure. Without loss of generality, ObjectRank assumes that each node has a tuple of attribute name/attribute value pairs. One may assume richer semantics by including the metadata of a node in the set of keywords. For example, the metadata "Forum", " Year", "Location" could be included in the keywords of a node.

Each edge e from u to v is labeled with its role $\lambda(e)$. For simplicity, we assume that there are no parallel edges and we will often denote an edge e from u to v as $u \to v$.

Fig. 4 shows the schema graph which is generated from Fig. 3. The schema graph $G(V_G, E_G)$ is a directed graph that describes the structure of D. Every node has an associated label. Each edge is labeled with a role. We say that a data graph $D(V_D, E_D)$ conforms to a schema graph $G(V_G, E_G)$ if there is a unique assignment μ such that:

1. For every node $v \in V_D$ there is a node $\mu(v) \in V_G$ such that $\lambda(v) = \lambda(\mu(v))$;

2. For every edge $e \in E_D$ from node u to node v there is an edge $\mu(e) \in E_G$ that goes from $\mu(u)$ to $\mu(v)$ and $\lambda(e) = \lambda(\mu(e))$.

From the schema graph $G(V_G, E_G)$, we create the authority transfer schema graph $\$G^A(V_G, E^A)$ to reflect the authority flow through the edges of the graph. This may be either a trial and error process, until we are satisfied with the quality of the results, or a domain expert's task. In particular, for each edge $e_G = u \to v$ of $E_G$, two authority transfer edges, $e_G^f = (u \to v)$ and $e_G^b = (v \to u)$ are created. The two edges carry the label of the schema graph edge and, in addition, each one is annotated with a (potentially different) authority transfer rate - $\alpha(e_G^f)$ and $\alpha(e_G^b)$ correspondingly. We say that a data graph conforms to an authority transfer schema graph if it conforms to the corresponding schema graph. (Notice that the authority transfer schema graph has all the information of the original schema graph.) Fig. 5 shows the authority transfer schema graph that corresponds to the schema graph in Fig. 4 (the edge labels are omitted). The motivation for defining two edges for each edge of the schema graph is that authority potentially flows in both directions and not only in the direction that appears in the schema. For example, a paper passes its authority to its authors and vice versa. Notice however, that the authority flow in each direction (defined by the authority transfer rate) may not be the same. For example, a paper that is cited by important papers is clearly important but citing important papers does not make a paper important.

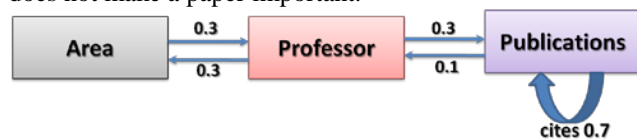

Fig . 5 The ObjectRank authority transfer schema graph.

Given a data graph $D(V_D, E_D)$ that conforms to an authority transfer schema graph $G^A(V_G, E^A)$, ObjectRank derives an authority transfer data graph $D^A(V_D, E_D^A)$ as follows.

For every edge $e = (u \rightarrow v) \in E_D$ the authority transfer data graph has two edges $e^f = (u \rightarrow v)$ and $e^b = (v \rightarrow u)$. The edges $e^f$ and $e^b$ are annotated with authority transfer rates $\alpha(e^f)$ and $\alpha(e^b)$. Assuming that $e^f$ is of type $e_G^f$, then

$$\alpha(e^f) = \begin{cases} \dfrac{\alpha(e_G^f)}{OutDeg(u, e_G^f)} & \text{if} \quad OutDeg(u, e_G^f) > 0 \\ 0 & \text{if} \quad OutDeg(u, e_G^f) = 0 \end{cases}$$

where $OutDeg(u, e_G^f)$ is the number of outgoing edges from u, of type $e_G^f$. The authority transfer rate $\alpha(e^b)$ is defined similarly. Fig. 6 illustrates the authority transfer data graph that corresponds to the data graph of Fig. 3 and the authority schema transfer graph of Fig. 4. Notice that the sum of authority transfer rates of the outgoing edges of a node u of type $\mu(u)$ may be less than the sum of authority transfer rates of the outgoing edges of $\mu(u)$ in the authority transfer schema graph, if u does not have all types of outgoing edges.

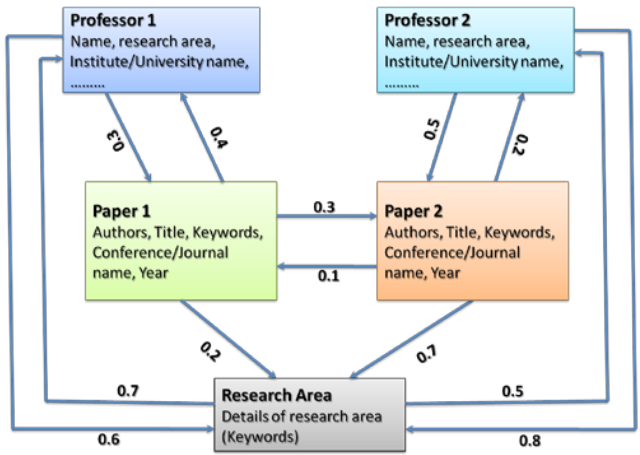

Fig. 6 Authority transfer data graph.

Then the total score of a page will be

$$r_G(v) = (1-d) + d \sum A_{ij}$$

where $A_{ij} = \alpha(e)$ if there is an edge $e = v_j \rightarrow v_i$ in $E_D^A$ and 0 otherwise, the damping factor d determines the portion of ObjectRank that an object transfers to its neighbors as opposed to keeping to itself. It was first introduced in the original PageRank technique [21], where it was used to ensure convergence in the case of PageRank sinks. The value for d used by PageRank is 0.85.

On the other hand, HITS is applicable in Semantic Web. The Semantic Web graph can be described by an adjacency matrix. For a network graph matrix M the well known authority ranking methods like HITS can be applied. HITS defines the authority ranking problem through mutual reinforcement between so-called hub and authority scores of graph nodes. The authority (relevance) score of each node is defined as the sum of hub scores of its predecessors. Analogously, the hub (connectivity) score of each node is defined as a sum of the authority scores of its successors.

The HITS team makes use of the relationship between hubs and authorities via an iterative algorithm works as follow: with each page A, they associate a non-negative authority weight $x_{hpi}$ and a non-negative hub weight $y_{hpi}$. They maintain the invariant that the weights of each type are normalized so their squares sum to 1:(Here also we need to take in our concern the type of the relations that exist between the SWDs we need to add this to the equations)

$$\sum_{A \in S_\sigma} (x^{\langle A \rangle})^2 = 1$$

$$\sum_{A \in S_\sigma} (y^{\langle A \rangle})^2 = 1$$

They view the pages with larger x and y-values as being "better" authorities and hubs respectively. If A points to many pages with large x-values, then it should receive a large y-value; and if A is pointed to by many pages with large y-values, then it should receive a large x-value. This motivates the definition of two operations on the weights, which denote by I and O. The I operation updates x-weights as follows:

$$x^{\langle A \rangle} \leftarrow \sum_{q:(q,A) \in E} y^{\langle q \rangle}$$

The O operation updates the y-weights as follows:

$$y^{\langle A \rangle} \leftarrow \sum_{q:(q,A) \in E} x^{\langle q \rangle}$$

Thus I and O are the basic means by which hubs and authorities reinforce one another. Therefore, to find the desired "equilibrium" values for the weights, one can apply the I and O operations in an alternating fashion, and see whether a fixed point is reached.

### 3.5 Metadata and Ontologies Repository

The metadata and ontologies repository is created to store the processed data for each SWD. The stored data can be used to derive analytical reports, such as classification of SWOs and SWDBs, rank of SWDs, and the IR index of SWDs.

### 3.6 User Interface

Every time a user submits a query, the proposed system analyzes it and tries to identify the ontological elements which are stored in the metadata and ontologies repository. Then, it is able to suggest to the user the potential meanings of his query that it recognized. The user is therefore presented with both the results of the syntactic search and the available meanings extracted from the query. This can help him to refine his request, disambiguating among its possible acceptations. When a user query is re-conducted to a specific meaning, the proposed system is able not only to look up resources semantically related to that meaning, but also to seek other concepts that could be of interest for the user. This is possible because the system can navigate across the sub-graph of interconnected elements of the domain ontology to generate the corresponding hubs and authorities.

## 4. Implementation And Results

For our experiments, we implemented our system in Java. The experiments were performed on a single PC with an Intel 1.73 GHZ Duo processor with 3GB RAM. We run an experiment to measure the effect of the total size of the sub-graph on the quality of the result. The total size of the sub-graph depends on two parameters. The first parameter is n which represents the number of pages that the sub-graph should start with. The second parameter is d which presents the number of pages a single page can bring into the sub-graph from the pages that pointing to it.

For our experiment we generate a comprehensive set of sub-graphs with 24 combinations of n and d. for each combination we measure the performance of our rank, i.e. the query time an quality of two lists.

Fig. 7 shows the effect of d on sub-graph construction time. Bigger d implies that more time to construct the sub-graph. Therefore, the quality of our rank algorithm is strongly affected by d. Thus, one has to strike balance between the quality of results and the time needed to construct the sub-graph.

## 5. Conclusion

Search engines are becoming such a powerful tools not only to find textual resources but also to analyze the contents of the document to get precise search result. Therefore, syntactic techniques are used to extract ontologies and metadata from the SWDs to calculate an accurate classification of the processed documents. The lexical and conceptual characteristics of a domain in an ontology are captured to prove that Semantic Web technologies provide real benefits to end users in terms of an easier and more effective access to information.

We developed an indexing and ranking technique which can be used in a Semantic Web search engine to facilitate the development of the Semantic Web and finding a proper ontology for the submitted search query. The entire documents are processed to extracted ontologies and find the relationships between the processed documents and the others. Therefore, our system is not based only on the extracted metadata from the document but also on extracted ontologies and the relations between documents. In other words, our main goals were to find a good measure for indexing the processed SWDs with extracting the proper ontologies, create a meaningful rank measure which reflects the importance of the processed document, and answer the user's queries efficiently.

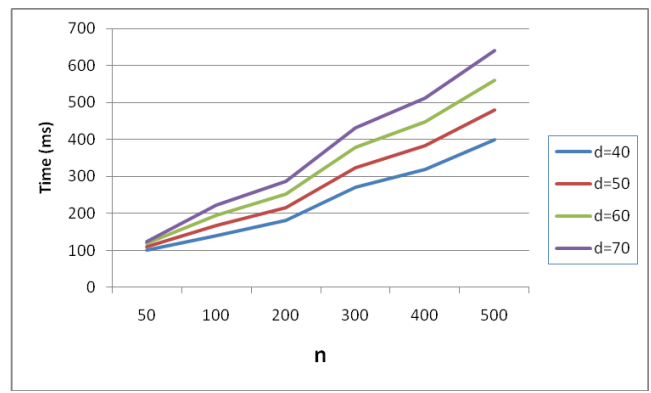

Fig. 7 The effect of n and d values on processing time.

**Ahmed Tolba** got his PhD degree in Electrical Engineering from Wuppertal University, Germany, in 1988, on Computer Vision. He is working as a Dean of the Faculty of Computer Studies at Arab Open University in Kuwait. He is also a professor in department of Computer Science, Faculty of Computers and Information, Mansoura University, Egypt. He published more than 100 previewed papers in international journals and conferences. He is interested in Artificial Intelligence, natural language processing, computer vision, and E-learning.

**Nabila Eladawi** got her B.Sc. in information systems from faculty of computers and information , Mansoura University, Egypt, in 2002. She is working as a demonstrator in the department of information systems, faculty of computers and information, Mansoura University, Egypt. She is interested in Semantic Web, Search engines, and natural language processing.

**Mohammed Elmogy** got his PhD degree in computer science from Hamburg University, Germany, in 2010, on Robotics. He is working as an assistant professor in information systems department, faculty of computers and information, Mansoura University, Egypt. He is interested in Robotics, Artificial Intelligence, Semantic Web, and Computer Vision.